\documentclass[10pt,twocolumn,letterpaper]{article}
\usepackage[utf8]{inputenc}
\usepackage{longtable}
\usepackage{iccv}
\usepackage{times}
\usepackage{epsfig}
\usepackage{graphicx}
\usepackage{amsmath}
\usepackage{amssymb}
\usepackage{caption}
\usepackage{subcaption}
\usepackage{gensymb}
\usepackage{textcomp}
\usepackage{pifont}
\newcommand{\xmark}{\ding{55}}
\usepackage[pagebackref=true,breaklinks=true,letterpaper=true,colorlinks,bookmarks=false,draft]{hyperref}

\iccvfinalcopy % *** Uncomment this line for the final submission

\begin{document}

%%%%%%%%% TITLE
% \title{Unsupervised Domain-Specific Deblurring with Scale-Adaptive Attention}
\title{Unsupervised Domain-Specific Deblurring using Scale-Specific Attention}
\author{Praveen Kandula\\
IIT Madras\\
% Institution1 address\\
% {\tt\small firstauthor@i1.org}
% For a paper whose authors are all at the same institution,
% omit the following lines up until the closing ``}''.
% Additional authors and addresses can be added with ``\and'',
% just like the second author.
% To save space, use either the email address or home page, not both
\and
Rajagopalan.A.N\\
IIT Madras\\
% First line of institution2 address\\
% {\tt\small secondauthor@i2.org}
}

\maketitle

\newcommand\run{\star}

%%%%%%%%% ABSTRACT
\begin{abstract}
In the literature, coarse-to-fine or scale-recurrent approach i.e. progressively restoring a clean image from its low-resolution versions has been successfully employed for single image deblurring. However, a major disadvantage of existing methods is the need for paired data; i.e. sharp-blur image pairs of the same scene, which is a complicated and cumbersome acquisition procedure. Additionally, due to strong supervision on loss functions, pre-trained models of such networks are strongly biased towards the blur experienced during training and tend to give sub-optimal performance when confronted by new blur kernels during inference time. To address the above issues, we propose unsupervised domain-specific deblurring using a scale-adaptive attention module (SAAM). Our network does not require supervised pairs for training, and the deblurring mechanism is primarily guided by adversarial loss, thus making our network suitable for a distribution of blur functions. Given a blurred input image, different resolutions of the same image are used in our model during training and SAAM allows for effective flow of information across the resolutions. For network training at a specific scale, SAAM attends to lower scale features as a function of the current scale. Different ablation studies show that our coarse-to-fine mechanism outperforms end-to-end unsupervised models and SAAM is able to attend better compared to attention models used in literature. Qualitative and quantitative comparisons (on no-reference metrics) show that our method outperforms prior unsupervised methods.
\end{abstract}
\begin{figure} 
    \centering
	\begin{subfigure}[b]{0.16\textwidth}
		\includegraphics[width=0.95\linewidth]{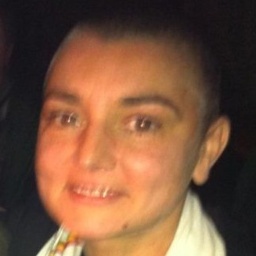}
		\caption{Blurred}
		\label{fig:introFig_inp}
	\end{subfigure}%
	\begin{subfigure}[b]{0.16\textwidth}
		\includegraphics[width=0.95\linewidth]{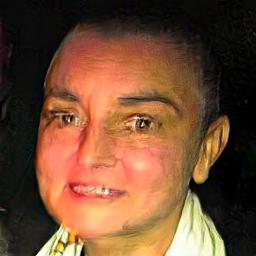}
		\caption{\cite{lu2019unsupervised}}
		\label{fig:introFig_sm}
	\end{subfigure}%
	\begin{subfigure}[b]{0.16\textwidth}
		\includegraphics[width=0.95\linewidth]{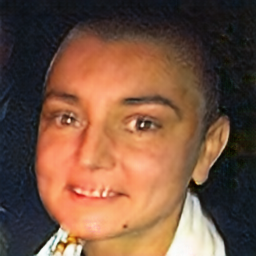}
		\caption{Ours}
		\label{fig:introFig_ours}
	\end{subfigure}	\\
		\begin{subfigure}[b]{0.16\textwidth}
		\includegraphics[width=0.95\linewidth]{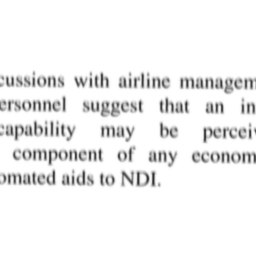}
		\caption{Blurred}
	\end{subfigure}%
	\begin{subfigure}[b]{0.16\textwidth}
		\includegraphics[width=0.95\linewidth]{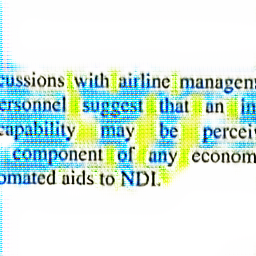}
		\caption{\cite{cyclegan}}
% 		\label{fig:introFig_sm}
	\end{subfigure}%
	\begin{subfigure}[b]{0.16\textwidth}
		\includegraphics[width=0.95\linewidth]{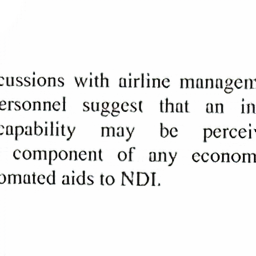}
		\caption{Ours}
% 		\label{fig:introFig_ours}
	\end{subfigure}%	
	\caption{Comparison of deblurring results on real blurred images with prior unuspervised mathods. (a) Blurred image from \cite{real_blur}, (b) result using pretrained model of \cite{lu2019unsupervised} and (c) Our result. (d) is the text image taken from \cite{text} and (e) is the result of \cite{cyclegan} retrained on text datset \cite{text}. (f) Our result.}
	\label{intro}

% 	and red line in (a,b,c) helps in quantifying rolling shutter rectification of \cite{rengarajan2017unrolling} and ours. The angles (between red and yellow lines) in these images are (a) 3.81\degree (b) 3.925\degree (c) 1.99\degree. In (b), \cite{rengarajan2017unrolling} increases the angle(3.81\degree to 3.925\degree), but (c) is able to reduce the original angle by almost 50\%. (\cite{rengarajan2017unrolling}).} 
\end{figure}
\section{Introduction}
Blur is an undesired phenomenon observed due to relative motion between camera and object of interest. Although blur can be used for aesthetic purposes, e.g. portrait images \cite{portrait}, bokeh effect \cite{bokeh} etc it adversely affects the performance of several computer vision applications like face recognition \cite{lu2019blurface}, object recognition \cite{kupyn2018deblurgan} and classification \cite{blurclassification}. Given a blurred image, deblurring aims to restore the underlying latent sharp image. 

Differnet algorithms \cite{deblurring1, deblurring2, deblurring3, deblurring4, vasu2018non, vasu2017local, deblurring7, deblurring8, deblurring_purohit1,purohit2020region} are proposed to tackle the problem of single image deblurring. Conventional methods pose this problem as one of estimating the underlying camera motion or blur kernel using an optimization framework. Due to the ill-posed nature of the problem, different assumptions \cite{kernel_sparcity, xu2013unnaturall0, yan2017imagebright, vasu2017local} were made on the image model and the nature of blur. Although these methods perform well on generic images, they cannot be generalized for domain-specific conditions such as faces \cite{faces} and text \cite{text}. Different priors have been proposed to handle domain-specific blur \cite{textdeblurring, facedeblurring}. However, these methods heavily depend on selection of priors, their weights and stoppage point during optimization, limiting their deblurring quality.

\begin{figure*}
	\centering
	\includegraphics[width=0.8\linewidth]{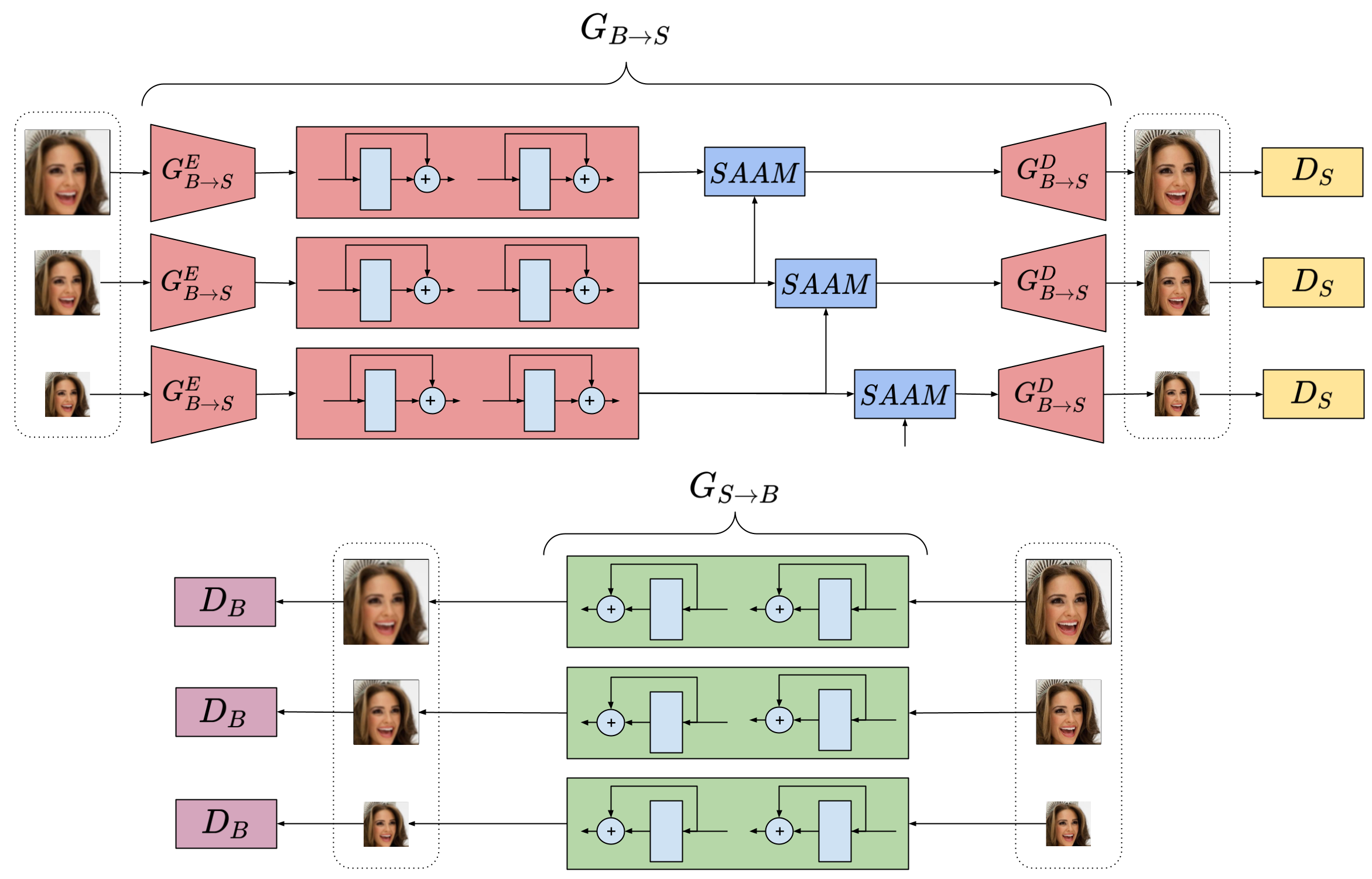}
	\caption{Proposed unsupervised scale adaptive attention deblurring network (USAAD). }
	\label{arch}
\end{figure*}

With the advent of deep learning, convolutional neural network (CNN) based supervised methods \cite{kupyn2018deblurgan, nah2017deep, faces, supervised4, purohit2020region, suin, nimisha2017blur, vasu2018non} were proposed for the task of deblurring. These algorithms preclude the necessity of assuming any priors due to implicit learning of weight parameters during training. However, the above methods require large amounts of paired training data which is cumbersome and complicated to obtain. Additionally, due to strong supervision 
of loss functions during training, these networks give sub-optimal performance when they encounter new types of blur during inference. 
 
Recently, a couple of unsupervised deblurring \cite{madam2018unsupervised, lu2019unsupervised} learning methods were proposed to relax the necessity of paired training data. Nimisha \textit{et al.} \cite{madam2018unsupervised} use a generative adversarial network (GAN) to transfer images from blur to sharp. They additionally use a reblurring network and gradient loss to maintain fidelity. Boyu \textit{et al.} \cite{lu2019unsupervised} proposed an unsupervised network where blur can be disentangled into the encoder network using KL divergence loss. These methods pose deblurring as an end-to-end problem where GAN loss used for training is calculated on the image at a single scale. As a result, these methods give suboptimal performance while handling coarse as well as fine-grained details.  

This paper addresses the above challenges by using a multi-scale architecture with a scale-adaptive attention module (SAAM). 
Several multi-scale deblurring algorithms have been proposed in the past that use a coarse-to-fine mechanism to take advantage of blur cues at different processing scales. These methods train the same network at different scales of the input image, resulting in a deblurring model that can handle both coarse and fine-level details. However, these methods require paired data and the supervised loss greatly aids in stability during training. Different from the above, we propose a multi-scale network for deblurring in an unsupervised setting. Training instability in GANs is well-studied in the literature, and several solutions \cite{dcgan} were proposed.In our approach, instead of cascading or directly adding features across different scales, SAAM attends to the feature maps of lower scales as a function of the present scale. The advantage of this mechanism is multi-fold. First, our structure's hidden states use information from different scales due to shared parameters, leading to better deblurring quality (see Fig. \ref{intro}) than prior unsupervised methods. Second, the multi-scale approach reduces the training instability problems such as mode collapse and unwarranted artifacts in the resultant image observed in GANs. Also, our SAAM module helps to select relevant information from lower scales, further improving the deblurring quality. 
% Instead of cascading or directly adding features across different scales, we propose a scale-dependent attention module (SAAM). SAAM is a self-attention module that attends feature maps on the lower scale as a function of the present scale.
Different ablation studies show that coarse-to-fine mechanism using SAAM gives better deblurring results than end-to-end counterparts devoid of recurrent connections.

.

Our contributions are summarized below:
\begin{itemize}
\item We propose an unsupervised deblurring network with multi-scale architecture and a scale-dependent attention module. Different ablation studies show that scale recurrent networks give superior performance compared to end-to-end methods in an unsupervised setting.
\item We further show that SAAM facilitates better information flow across different scales in contrast to directly cascading or adding feature maps. We further show the efficacy of using SAAM over different attention modules.

\item We provide extensive comparisons on supervised and unsupervised methods and show that our method performs favourably against supervised and outperforms unsupervised methods qualitatively and quantitatively (on no-reference metrics) when tested on different datasets.
\end{itemize}

%-------------------------------------------------------------------------
\section{Related works }
Image deblurring is an active research area in the vision community for the past two decades. Many algorithms \cite{video1, video2} were proposed for video deblurring, where multiple frames for the same scene are available, making it easier to estimate camera motion. However, the problem becomes ill-posed when only a single image is available, a common occurrence. Many priors were assumed on the latent sharp image and nature of camera motion to solve this problem. Some of these include dark channel prior \cite{darkchannel1}, bright channel prior \cite{yan2017imagebright}, heavily tailored gradient prior \cite{xu2013unnaturall0}. These priors are used to estimate the underlying latent image and camera motion using alternating minimization techniques. Although these methods do not require any paired training data, the deblurring results heavily depend on the choice of prior and fail to generalize to other blur models.
\begin{figure*}
	\centering
	\includegraphics[width=1.0\linewidth]{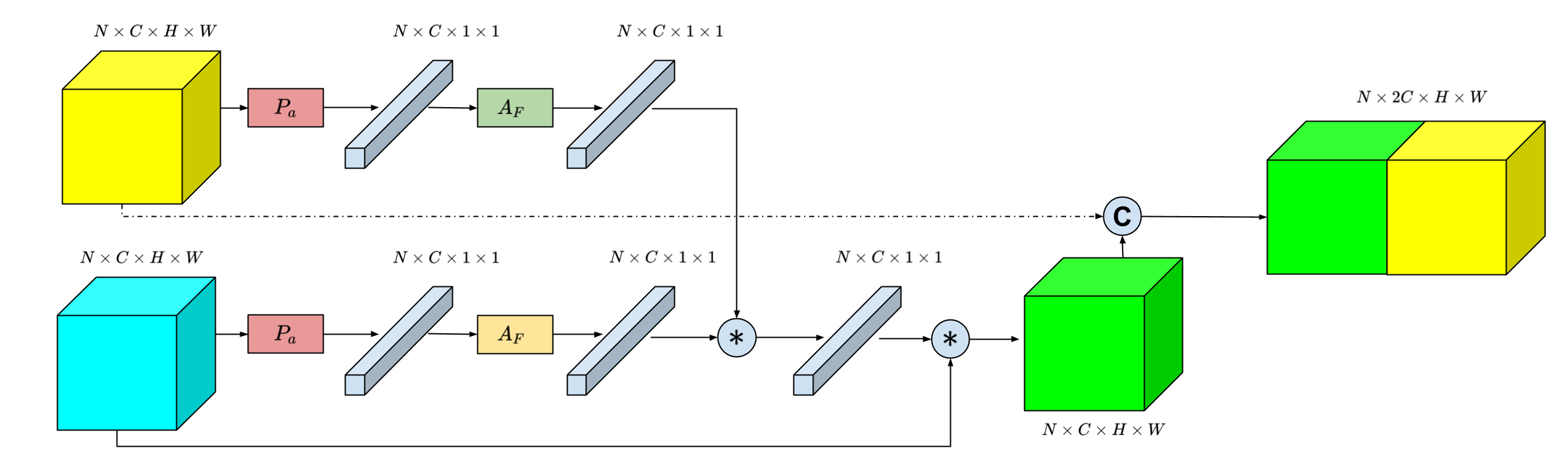}
	\caption{Proposed scale adaptive attention module (SAAM). The cyan block ($N\times C\times H \times W$) is the input feature map from the lower resolution layer, while the yellow block ($N\times C\times H \times W$) is the feature map from the higher resolution layers. $P_a$ refers to the average polling layer, and $A_F$ is convolutional network block. Note that there are two independent $A_F$'s each operating on the feature maps of its respective resolution layers (best viewed in color).}
	\label{SAAMDetail}
\end{figure*}
Due to recent advances in deep learning, many algorithms are proposed to estimate blur kernel and sharp image using neural networks instead of optimization methods. Sun \textit{et al.} \cite{blurmodel2} proposed a CNN based network to estimate blur kernel, while \cite{frequency} uses CNN to estimate Fourier coefficients for deblurring in the frequency domain. Unlike the above, Nah \textit{et al.} \cite{nah2017deep} proposed an end-to-end network for direct estimation of sharp images without estimation of the blur kernel. Kupyn \textit{et al.} \cite{kupyn2018deblurgan} propose deblur GAN where an adversarial loss is used along with supervised loss. Although these methods give good deblurring results, a significant disadvantage is the requirement of paired training data. While different blurring \cite{kupyn2018deblurgan, blurmodel2, frequency} schemes were proposed to circumvent this problem,  these supervised networks are heavily biased towards the blur used during training and give sub-optimal performance when faced with new blur kernels. 

\textbf{Domain-specific methods}
The generic priors designed for natural images are not appropriate for domain-specific images. There is a drop in the performance when these methods are used to restore specific cases such as faces or text. Several restoration methods \cite{domian3, chrysos2017deepdomain1, rengarajan2017unrollingdomain2, text, anwar2015classdomain4, textdeblurring, facedeblurring} were proposed to counter the inability of the above methods while tackling domain-specific images. Pan \textit{et al.} \cite{facedeblurring} extract important structures from a set of exemplar faces and use them to guide the deblurring process. Hradis \textit{et al.} \cite{text} train an end-to-end deblurring network for text deblurring and show improvement in OCR. Pan \textit{et al.} \cite{textdeblurring} proposed an $L_0$ regularized intensity and gradient prior using half-quadratic splitting for text deblurring.

\textbf{Scale recurrent methods} Of late, there is an increased interest in using scale recurrent and coarse-to-fine generation schemes. Scale recurrent methods have been shown to yield state of the art results for image restoration tasks. \cite{nah2017deep} trains a network at different scales for the task of deblurring. The lower scale output acts as an input to the next higher scale along with the input blur image. Instead of regressing for the clean image at every scale, \cite{zhang2019deep} proposed to regress for the clean image only at the highest scale, and the residual in the lower scales are added to the next scale. \cite{suin2020spatially} proposed a multi-scale architecture where the filter weights and receptive field change according to the input image. The coarse-to-fine mechanism is recently used in unsupervised algorithms as it can provide increased stability for adversarial training. A high dimensional image synthesis network is proposed in \cite{karras2020analyzing} using a coarse-to-fine image generation scheme staring from 4x4 extending all the way to 1024. \cite{shaham2019singan} proposed a scale recurrent network for generating different instances from a single image using adversarial loss.

\section{Proposed method}
Our proposed network, unsupervised scale adaptive attention deblurring network (USAAD), is illustrated in Fig. \ref{arch}, along with the scale-adaptive attention module (SAAM) in Fig. \ref{SAAMDetail}. Our network architecture is inspired by the recent success of scale recurrent structures in image restoration tasks. Given a blurred image $I^b_{M}$, three samples of input image are used for training i.e., $I^b_{M}$, $I^b_{M/2}$ and $I^b_{M/4}$ where $I^b_{p}$ denotes input image downsampled to $pxp$ dimension. The training mechanism of our algorithm has three steps for every input image. First, at the coarsest scale, generator \text{$ G_{B \rightarrow S}$} converts $I^{b}_{M/4}$ from blur to sharp  domain using adversarial loss. \text{$ G_{B \rightarrow S}$} consists of three networks, a encoder network \text{$ G_{B \rightarrow S}^E$}, followed by a series of nine residual blocks \cite{residual} and decoder network \text{$ G_{B \rightarrow S}^D$}. \text{$ G_{S \rightarrow B}$} blurs the generated sharp image which is then compared with the input image to maintain fidelity of contents. The same procedure is followed in the next scale with  $I^b_{M/2}$, except that the decoder, \text{$ G_{B \rightarrow S}^{D} $} , takes the output of SAAM instead of the final residual block. SAAM helps the present scale to use important information from the previous scale to improve deblurring quality (see Sec. 3.1). The same procedure is repeated at the finest scale with $I^b_{M}$, and the estimated sharp image is the final restored output. The deblurring mechanism of our method can be represented as 
\begin{equation*}
    \mathcal{I}^i, \mathcal{F}^i = Net_{USAAD} ( \mathcal{I}^{i-1}, \mathcal{F}^{i-1}, \mathcal{B}^i;\theta_{USAAD})
\end{equation*}
where $i$ denotes the present scale and $i\in{1,2,3}$. Inspired by \cite{nah2017deep}, we use three resolutions of the input image to train the network and $M = 256$ unless mentioned otherwise. $\mathcal{I}$, $\mathcal{F}$ and $\mathcal{B}$ denote estimated sharp image, output features of last residual block and input blurry image, respectively, and $\theta$ denotes learnable parameters of our network. The generator and discriminator networks in our architecture share the same parameters.
% Although supervised methods \cite{zhang2019deep, suin2020spatially} regress only for residual at each scale, our unsupervised network regress for sharper images to counter training instability due to GANs. 
% Along with the sharing parameters of netowrks across resolutions, the information across different resolutions is passed effectively using SAAM to improve deblurring quality. 
The following subsections give detailed discussion of the SAAM module followed by loss functions used in our model and network architecture.

\subsection{Scale-adaptive attention module (SAAM)}

The objective of SAAM is to use information from the previous scale to improve the deblurring quality at the present scale. A trivial way to achieve this is to directly concatenate or add features from the last residual blocks of both the scales and pass them to the decoder. However, not all the lower-scale features are equally important in improving the deblurring quality. Therefore, concatenating or adding the entire set of lower-scale features can result in sub-optimal performance due to irrelevant channels.  Instead of considering each channel equally, SAAM uses both the lower and higher scale feature maps to selectively pay attention to more relevant channels in the lower-scale features. Similar to channel attention \cite{chen2017sca}, SAAM can be seen as a process of selecting relevant semantic attributes. 
% \begin{align*}
%     V = \text{Concat}(U^a_{X}, U_{2X})
% \end{align*}
\begin{figure*}[t!]
% \hskip-0.7cm
  \centering
  \begin{subfigure}[b]{0.09\linewidth}
    \includegraphics[width=\linewidth]{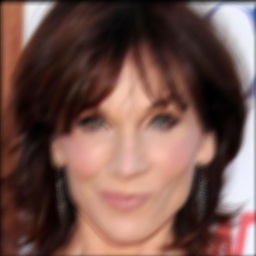}
    % \caption{}
  \end{subfigure}
  \begin{subfigure}[b]{0.09\linewidth}
    \includegraphics[width=\linewidth]{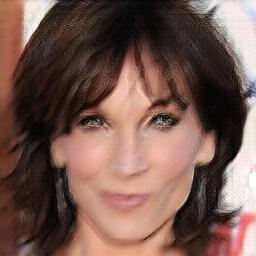}
    % \caption{}
    \end{subfigure}
  \begin{subfigure}[b]{0.09\linewidth}
    \includegraphics[width=\linewidth]{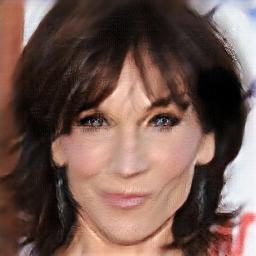}
    % \caption{}
    \end{subfigure}
   \begin{subfigure}[b]{0.09\linewidth}
    \includegraphics[width=\linewidth]{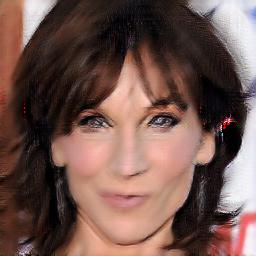}
    % \caption{}
    \end{subfigure}
  \begin{subfigure}[b]{0.09\linewidth}
    \includegraphics[width=\linewidth]{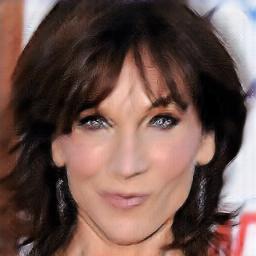}
    % \caption{}
  \end{subfigure}
  \begin{subfigure}[b]{0.09\linewidth}
    \includegraphics[width=\linewidth]{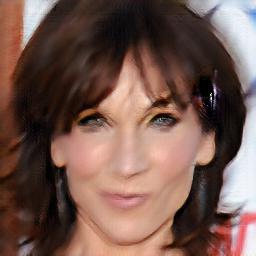}
    % \caption{}
  \end{subfigure}
  \begin{subfigure}[b]{0.09\linewidth}
    \includegraphics[width=\linewidth]{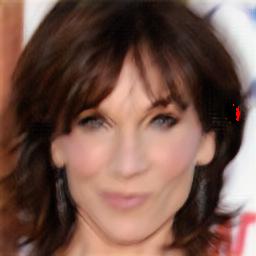}
    % \caption{}
  \end{subfigure}
  \begin{subfigure}[b]{0.09\linewidth}
    \includegraphics[width=\linewidth]{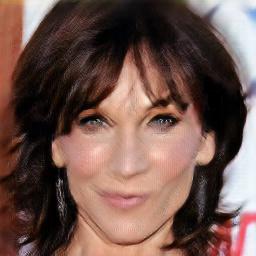}
    % \caption{}
  \end{subfigure}
  \begin{subfigure}[b]{0.09\linewidth}
    \includegraphics[width=\linewidth]{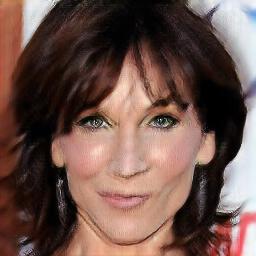}
    % \caption{}
  \end{subfigure}
  \begin{subfigure}[b]{0.09\linewidth}
    \includegraphics[width=\linewidth]{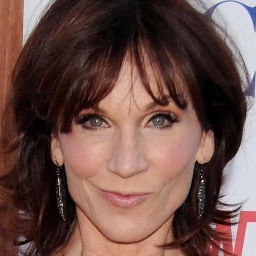}
    % \caption{}
  \end{subfigure}\\
 \begin{subfigure}[b]{0.09\linewidth}
    \includegraphics[width=\linewidth]{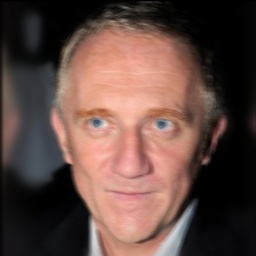}
    \caption{}
  \end{subfigure}
  \begin{subfigure}[b]{0.09\linewidth}
    \includegraphics[width=\linewidth]{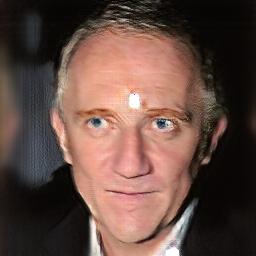}
    \caption{}
    \end{subfigure}
  \begin{subfigure}[b]{0.09\linewidth}
    \includegraphics[width=\linewidth]{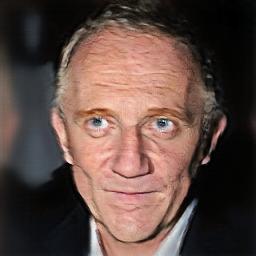}
    \caption{}
    \end{subfigure}
   \begin{subfigure}[b]{0.09\linewidth}
    \includegraphics[width=\linewidth]{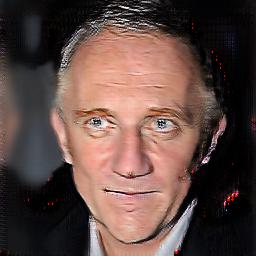}
    \caption{}
    \end{subfigure}
  \begin{subfigure}[b]{0.09\linewidth}
    \includegraphics[width=\linewidth]{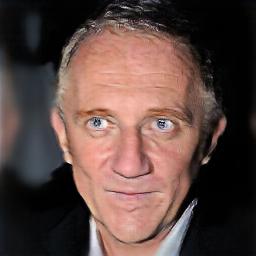}
    \caption{}
  \end{subfigure}
  \begin{subfigure}[b]{0.09\linewidth}
    \includegraphics[width=\linewidth]{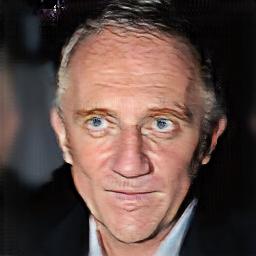}
    \caption{}
  \end{subfigure}
  \begin{subfigure}[b]{0.09\linewidth}
    \includegraphics[width=\linewidth]{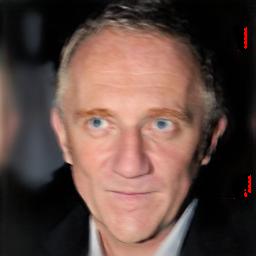}
    \caption{}
  \end{subfigure}
  \begin{subfigure}[b]{0.09\linewidth}
    \includegraphics[width=\linewidth]{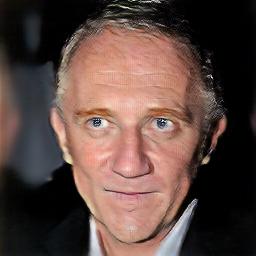}
    \caption{}
  \end{subfigure}
  \begin{subfigure}[b]{0.09\linewidth}
    \includegraphics[width=\linewidth]{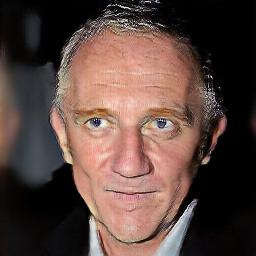}
    \caption{}
  \end{subfigure}
  \begin{subfigure}[b]{0.09\linewidth}
    \includegraphics[width=\linewidth]{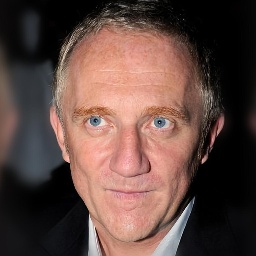}
    \caption{}
  \end{subfigure}
  \caption{Ablation study. (a) input blurry image and (j) is the sharp image. (b-i) are the resultant images of Net1-Net8. See section 4.2 for detailed explanation.}
  \label{fig:Ablation_study1}
\end{figure*}
SAAM takes two feature maps $U_{2X}, U_X \in \mathbb{R}^{N\times C\times H \times W}$, from the last residual block in  \text{$ G_{B \rightarrow S} $} at the present scale and the immediate previous scale, respectively. Here $N$ denotes the batch size, $C$ is the total number of channels and $H$, $W$ are the height and width of the feature map, respectively. Without loss of generality, we consider $N=1$ and we represent both the input feature maps as $U_{\run} = [u_{\run}^1, u_{\run}^2, ..., u_{\run}^C]$, where $u_{\run}^i \in \mathbb{R}^{H\times W}$ for $\run \in \{2X, X\}$. We apply mean pooling $(P_a)$ for each channel and get channel vectors for both the feature maps as

\begin{align*}
    u^M_{2X} &= [\bar{u}_{2X}^1,\bar{u}_{2X}^2,...,\bar{u}_{2X}^C] &\in \mathbb{R}^{C} \\
    u^M_{X} &= [\bar{u}_{X}^1,\bar{u}_{X}^2,...,\bar{u}_{X}^C] &\in \mathbb{R}^{C} 
\end{align*}
where $\bar{u}_{\run}^i$ is the mean of channel $u_{\run}^i$ features. The channel vectors $u^M_{2X}, u^M_{X}$ are passed through convolutional network $\Phi_{2X}$ and $\Phi_{X}$, respectively (denoted as $A_F$ in Fig. \ref{SAAMDetail}), to obtain the learned scale attention representations $v_{2X}, v_{X}$ where

\begin{align*}
    v_{\run} = \Phi_{\run}(u^M_{*}) \ \ \run \in \{2X, X\}
\end{align*}

The effective channel attention vector $\beta \in \mathbb{R}^C$ is defined as a function of $v_{2X}$ and $v_{X}$ as follows, 

\begin{align*}
    \beta = \sigma\left(v_{2x} \times v_{X}\right) \in \mathbb{R}^C
\end{align*}
where $\sigma$ denotes the sigmoid function, and $\times$ refers to element wise multiplication. Sigmoid activation is used to normalize the attention weights between 0 and 1 to represent the channel importance. The multiplication of scale attention representations $(v_{\run}'s)$ ensures that the channel representations which are aligned get greater attention than misaligned channels.

Channel attention is applied on $U_{X}$ by multiplying channel-wise the attention coefficients $\beta$ , which can be represented as $U^a_{X} \in \mathbb{R}^{N\times C\times H\times W}$,

\begin{align*}
    U^a_{X} = \beta \odot U_X 
\end{align*}
where $\odot$ refers to channel-wise multiplication. The resultant lower scale feature map is concatenated with the higher resolution feature map $U_{2X}$ along the channel dimension and passed through the decoder. 
% and is returned as output $V \in \mathbb{R}^{N\times 2C \times H \times W}$. 
This procedure ensures that lower scale feature information relevant for deblurring is effectively passed on to higher resolution layers.

\subsection{Loss functions}
Given a real blur image ($I^b$), the generator network \text{$ G_{B \rightarrow S} $} transfers the image from blur to sharp domain. The output $\hat{I}^s$ of decoder \text{$ G_{B \rightarrow S}^{D} $} is used by discriminator \text{$ D_s$} to distinguish if the resultant image is sharp or not. 
\begin{equation*}
    \hat{I}^s = \text{$ G_{B \rightarrow S} $} (I^b)
\end{equation*}
The following loss function is used to optimize both generator \text{$ G_{B \rightarrow S} $} and discriminator  \text{$ D_s$} simultaneously

\begin{equation}
\begin{split}
 \mathbb{L}_{GAN}(G_{B \rightarrow S}, D_S) =  \mathbb{E}_{I_s \sim p(I_s)} [\log D_S(I_s)] +\\ \mathbb{E}_{I_b\sim p(I_b)} [\log(1 - D_S(G_{B \rightarrow S}(I_b)))]  
\end{split}
\end{equation}
where $\mathbb{E}$ is the error function, $p$ denotes the data distribution, $I_b\sim p(I_b)$ and $I_s\sim p(I_s)$ denote images sampled from blur and sharp image distributions respectively.

Akin to Eq. 1, the output of decoder \text{$ G_{S \rightarrow B}^{D} $} is used by discriminator \text{$ D_B$} to distinguish if the resultant image is blurred  or not. The loss function used to optimize both generator \text{$ G_{S \rightarrow B} $} and discriminator  \text{$ D_B$} simultaneously is

\begin{equation}
\begin{split}
 \mathbb{L}_{GAN}(G_{S \rightarrow B}, D_B) =  \mathbb{E}_{I_b \sim p(I_b)} [\log D_B(I_b)] +\\ \mathbb{E}_{I_s\sim p(I_s)} [\log(1 - D_B(G_{S \rightarrow B}(I_s)))]  
\end{split}
\end{equation}

The above adversarial loss functions are sufficient to generate visually sharp images. However, the estimated sharp image's content need not exactly match that the input image due to unavailability of supervised pairs. Inspired by cycleGAN \cite{cyclegan}, we use cycle consistency loss, where the estimated sharp image is projected into blur domain using \text{$ G_{S \rightarrow B} $}
and compared with the input blur image. The projected blur image can be represented as  

\begin{equation*}
    \hat{I}^b = \text{$ G_{S \rightarrow B} $} (I^s)
\end{equation*}
The cycle consistency loss function can be defined as 
\begin{equation}
\begin{split}
 \mathbb{L}_{cyc\_b}(G_{B \rightarrow S}, G_{S \rightarrow B}) = 
 \mathbb{E}_{I_b \sim p(I_b)}  \\
 [||G_{S \rightarrow B}(G_{B \rightarrow S}(I_b)) - I_b||_1] 
\end{split}
\end{equation}

Similarly, the cycle consistency loss can be applied for the other domain by projecting the estimated blur image to the sharp domain using \text{$ G_{B \rightarrow S} $} and comparing with the real sharp image. The resultant loss function can be defined as 
\begin{equation}
\begin{split}
 \mathbb{L}_{cyc\_s}(G_{S \rightarrow B}, G_{B \rightarrow S}) = 
 \mathbb{E}_{I_s \sim p(I_s)}  \\
 [||G_{B \rightarrow S}(G_{S \rightarrow B}(I_s)) - I_s||_1] 
\end{split}
\end{equation}

These loss functions are calculated at a single scale; however, since our network is trained for $n$ scales, the total loss function can be written as 

\begin{equation}
\begin{split}
  \mathbb{L}_{Total}(G_{S \rightarrow B}, G_{B \rightarrow S}, 
    D_S, D_B)  \\ =  \sum_{i=1}^{n} \lambda_{adv}\mathbb{L}_{GAN}^{i}(G_{B \rightarrow S}, D_S) \\ + \lambda_{adv}\mathbb{L}_{GAN}^{i}(G_{S \rightarrow B}, D_B) \\+ \lambda_{cyc}\mathbb{L}_{cyc\_s}^{i}(G_{S \rightarrow B}, G_{B \rightarrow S})   \\+ \lambda_{cyc} \mathbb{L}_{cyc\_b}^{i}(G_{B \rightarrow S}, G_{S \rightarrow B})
\end{split}
\end{equation}
where $n$ is the number of scales the network is trained on. We used $n = 3$ for our model following \cite{nah2017deep}. Following \cite{cyclegan}, the weights for $\lambda_{adv}$ and $\lambda_{cyc}$ are set as 1 and 10 respectively. The whole network is trained in a min-max fashion as 
\begin{equation*}
    \arg \underset{G_{S \rightarrow B}, G_{B \rightarrow S}}{\min}\;\underset{D_B, D_B}{\max}\; \mathbb{L}_{Total}(G_{S \rightarrow B}, G_{B \rightarrow S}, 
    D_S, D_B)
\end{equation*}
% \begin{align*}
%     V = \text{Concat}(U^a_{X}, U_{2X})
% \end{align*}
\begin{figure*}[t!]
% \hskip-0.7cm
\hskip-0.63cm
  \centering
  \begin{subfigure}[b]{0.09\linewidth}
    \includegraphics[width=\linewidth]{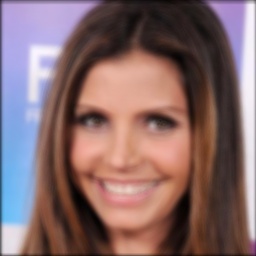}
  \end{subfigure}
  \begin{subfigure}[b]{0.09\linewidth}
    \includegraphics[width=\linewidth]{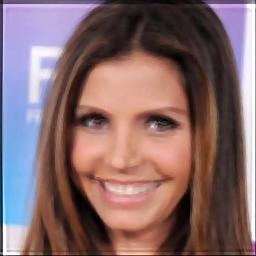}
  \end{subfigure}
  \begin{subfigure}[b]{0.09\linewidth}
    \includegraphics[width=\linewidth]{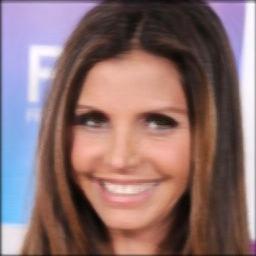}
  \end{subfigure}
  \begin{subfigure}[b]{0.09\linewidth}
    \includegraphics[width=\linewidth]{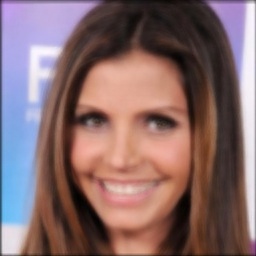}
  \end{subfigure}
  \begin{subfigure}[b]{0.09\linewidth}
    \includegraphics[width=\linewidth]{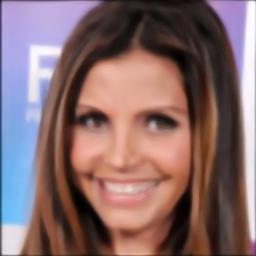}
  \end{subfigure}
  \begin{subfigure}[b]{0.09\linewidth}
    \includegraphics[width=\linewidth]{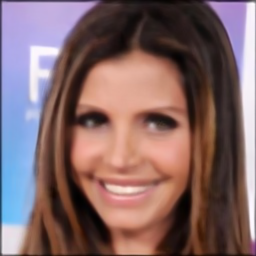}
  \end{subfigure}
  \begin{subfigure}[b]{0.09\linewidth}
    \includegraphics[width=\linewidth]{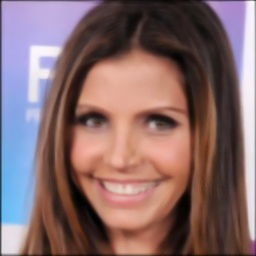}
  \end{subfigure} 
  \begin{subfigure}[b]{0.09\linewidth}
    \includegraphics[width=\linewidth]{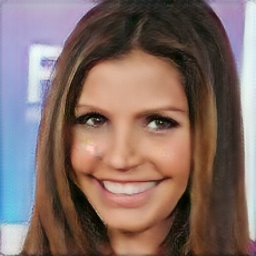}
  \end{subfigure}
  \begin{subfigure}[b]{0.09\linewidth}
    \includegraphics[width=\linewidth]{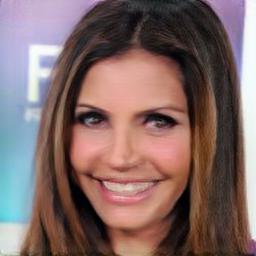}
  \end{subfigure}
  \begin{subfigure}[b]{0.09\linewidth}
    \includegraphics[width=\linewidth]{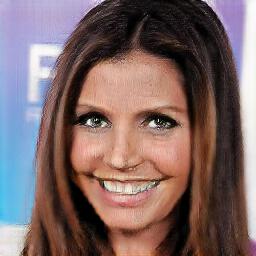}
  \end{subfigure}
  \begin{subfigure}[b]{0.09\linewidth}
    \includegraphics[width=\linewidth]{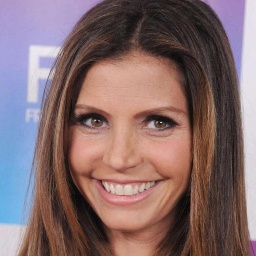}
    \end{subfigure}\\
     \hskip-0.63cm
 \begin{subfigure}[b]{0.09\linewidth}
    \includegraphics[width=\linewidth]{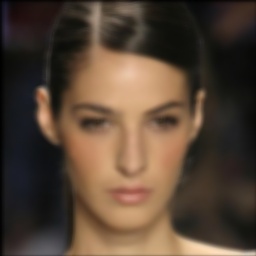}
    \caption{Blurred}
  \end{subfigure}
  \begin{subfigure}[b]{0.09\linewidth}
    \includegraphics[width=\linewidth]{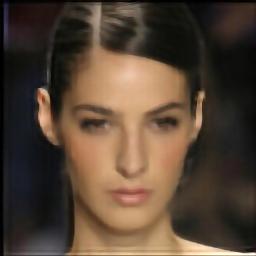}
    \caption{\cite{xu2013unnaturall0}}
    \end{subfigure}
  \begin{subfigure}[b]{0.09\linewidth}
    \includegraphics[width=\linewidth]{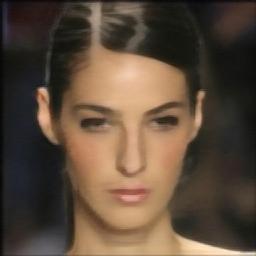}
    \caption{\cite{kupyn2018deblurgan}}
    \end{subfigure}
   \begin{subfigure}[b]{0.09\linewidth}
    \includegraphics[width=\linewidth]{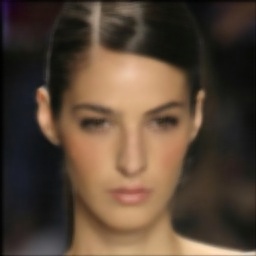}
    \caption{\cite{deblurganv2}}
    \end{subfigure}
  \begin{subfigure}[b]{0.09\linewidth}
    \includegraphics[width=\linewidth]{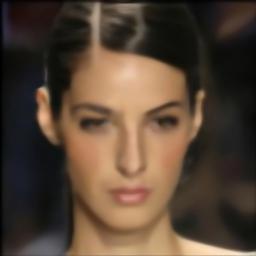}
    \caption{\cite{zhang2019deep}}
  \end{subfigure}
  \begin{subfigure}[b]{0.09\linewidth}
    \includegraphics[width=\linewidth]{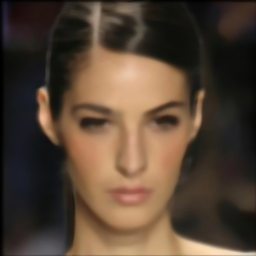}
    \caption{\cite{nah2017deep}}
  \end{subfigure}
  \begin{subfigure}[b]{0.09\linewidth}
    \includegraphics[width=\linewidth]{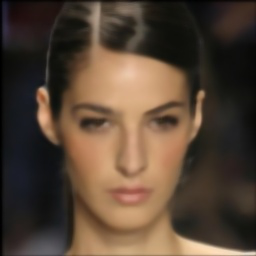}
    \caption{\cite{suin2020spatially}}
  \end{subfigure}
  \begin{subfigure}[b]{0.09\linewidth}
    \includegraphics[width=\linewidth]{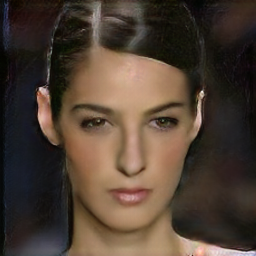}
    \caption{\cite{cyclegan}}
  \end{subfigure}
  \begin{subfigure}[b]{0.09\linewidth}
    \includegraphics[width=\linewidth]{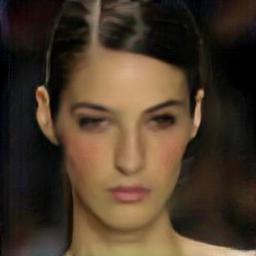}
    \caption{\cite{lu2019unsupervised}}
  \end{subfigure}
  \begin{subfigure}[b]{0.09\linewidth}
    \includegraphics[width=\linewidth]{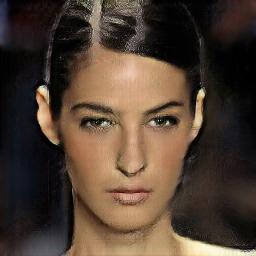}
    \caption{Ours}
  \end{subfigure}
  \begin{subfigure}[b]{0.09\linewidth}
    \includegraphics[width=\linewidth]{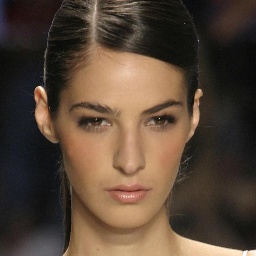}
    \caption{Sharp}
  \end{subfigure}
  \caption{Visual comparisons with start of the art results on face test dataset \cite{CelebAHQ}.}
  \label{fig:Visualcompar_face}
\end{figure*}

% \begin{align*}
%     V = \text{Concat}(U^a_{X}, U_{2X})
% \end{align*}

\subsection{Network architecture}
 The encoder network \text{$ G_{B \rightarrow S}^{E} $} in  Fig. \ref{arch}, consists of two convolutional layers with stride two, thus downsampling the input sample by a factor of four. A series of nine residual blocks follow the encoder network. At the coarsest level, the network cannot take features from the previous scale. However, the last residual block features are concatenated with the next level using the SAAM module. Finally, the concatenated features are passed through a decoder network, \text{$ G_{B \rightarrow S}^{D} $}, a mirror representation of the encoder, but deconvolutional layers replace the convolutional layers. The decoder's output is passed through \text{$ G_{S \rightarrow B} $}, which transfers the image from sharp to blur domain. \text{$ G_{S \rightarrow B} $} is a lightweight network with four convolutional layers using a filter size of 3 and maintaining the same spatial size using padding. Our reason for the simple architecture for \text{$ G_{S \rightarrow B} $} is to reduce the number of parameters and computational time. Also, the deblurring task is far more complicated than inducing blur into a sharp image. For discriminators $D_S$ and $D_B$, we use PatchGAN \cite{pix2pix} to differentiate between real and fake samples.

\section{Experiments}
This section is arranged as follows 1. Dataset creation and metrics used, 2. Ablation studies, 3. Comparisons on face and text testsets and 4. Visual comparisons on real face dataset.

\subsection{Dataset and metrics:}
\textbf{CelebA dataset:} We use the face dataset of \cite{CelebAHQ} to train our model. \cite{CelebAHQ} contains 30K face images and 700 images randomly selected and used as a test dataset for comparisons with state of the art methods. The remaining 29.3K images are grouped into two halves, and the blur model of \cite{kupyn2018deblurgan} is applied to one of the groups keeping the other intact. Thus unsupervised pairs of clean and blur face images are created for training.

\textbf{Text dataset:} We used text dataset provided by Hradis \textit{et al.} \cite{text} which contains large collection of 66K blur text images generated using motion and defocus blur. The 66K images are grouped into two halves, with one group containing the sharp images, while the other contains only blur images. The dataset is created such that there is no correspondence between the two groups. Since the images are already blurred, we did not apply any blur model, and the above dataset is used for training. We used a separate test dataset provided by \cite{text} to compare with competing methods.

\begin{table}[!htb]
    % \caption{Global caption}
    % \begin{minipage}{.5\linewidth}
      \centering

\caption{Quantitative comparisons of different ablation studies of our model on the face dataset. Scales indicate the number of resolutions the network was trained. $A.F$ and $C.F$ indicate that feature maps across the resolution are added and concatenated respectively, while $C.A$ and $S.A$ indicate channel \cite{channelattention} and spatial attention\cite{spatialattention} respectively.}

% \begin{tabular}{p{6cm}p{6cm}}
\hskip-0.65cm
\begin{tabular}{c|c|c|c|c|c|c|c}
    
        \hline
        Design & Scales & \textit{A.F} & \textit{C.F} & \textit{C.A} & \textit{S.A} & SAAM & brisque \\
        \hline
        Net1 &  1 & \xmark & \xmark &\xmark &\xmark & \xmark & 32.89 \\
        \hline
        Net2 &  2 & \xmark & \xmark &\xmark &\xmark & \xmark &31.29 \\
        \hline
        Net3 &  3 & \xmark & \xmark &\xmark &\xmark & \xmark &30.34 \\
        \hline
        Net4 &  3 & \checkmark & \xmark &\xmark &\xmark & \xmark &33.53 \\
        \hline
        Net5 &  3 & \xmark & \checkmark &\xmark &\xmark & \xmark &30.21 \\
        \hline
        Net6 &  3 & \xmark & \checkmark &\checkmark &\xmark & \xmark &29.52 \\
        \hline
        Net7 &  3 & \xmark & \checkmark &\xmark &\checkmark & \xmark &27.38 \\
        \hline
        Net8 &  3 & \xmark & \checkmark &\xmark &\xmark & \checkmark &\textbf{25.52} \\
        \hline
        
\end{tabular}
% \end{tabular}
\label{ablationTable}

\end{table}

\begin{figure*}[t!]

  \centering
  \begin{subfigure}[b]{0.09\linewidth}
    \includegraphics[width=\linewidth]{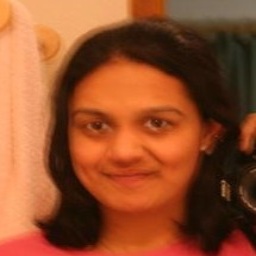}
  \end{subfigure}
  \begin{subfigure}[b]{0.09\linewidth}
    \includegraphics[width=\linewidth]{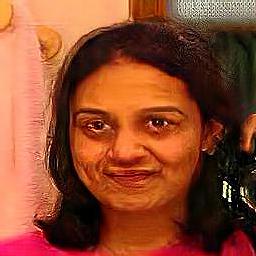}
  \end{subfigure}
  \begin{subfigure}[b]{0.09\linewidth}
    \includegraphics[width=\linewidth]{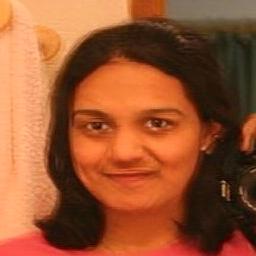}
  \end{subfigure}
  \begin{subfigure}[b]{0.09\linewidth}
    \includegraphics[width=\linewidth]{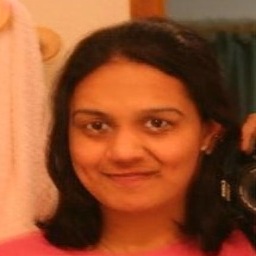}
  \end{subfigure}
  \begin{subfigure}[b]{0.09\linewidth}
    \includegraphics[width=\linewidth]{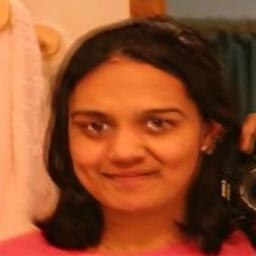}
  \end{subfigure}
  \begin{subfigure}[b]{0.09\linewidth}
    \includegraphics[width=\linewidth]{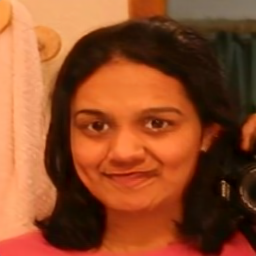}
  \end{subfigure}
  \begin{subfigure}[b]{0.09\linewidth}
    \includegraphics[width=\linewidth]{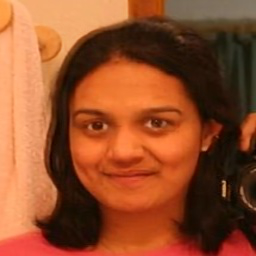}
  \end{subfigure} 
  \begin{subfigure}[b]{0.09\linewidth}
    \includegraphics[width=\linewidth]{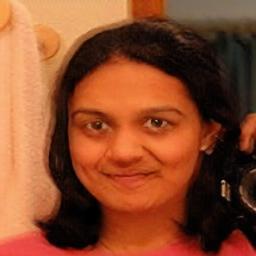}
  \end{subfigure}
  \begin{subfigure}[b]{0.09\linewidth}
    \includegraphics[width=\linewidth]{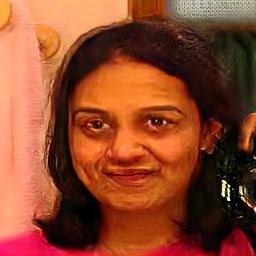}
  \end{subfigure}
  \begin{subfigure}[b]{0.09\linewidth}
    \includegraphics[width=\linewidth]{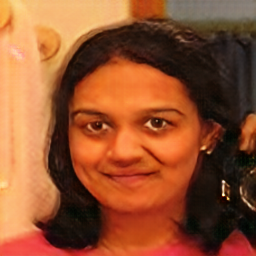}
  \end{subfigure} \\
 
 \begin{subfigure}[b]{0.09\linewidth}
    \includegraphics[width=\linewidth]{Comparisons/input/face.jpg}
    \caption{Blurred}
  \end{subfigure}
  \begin{subfigure}[b]{0.09\linewidth}
    \includegraphics[width=\linewidth]{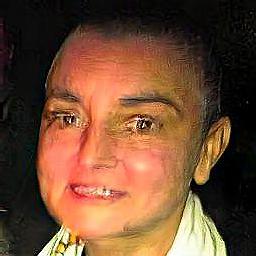}
    \caption{\cite{xu2013unnaturall0}}
    \end{subfigure}
  \begin{subfigure}[b]{0.09\linewidth}
    \includegraphics[width=\linewidth]{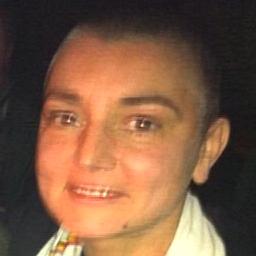}
    \caption{\cite{kupyn2018deblurgan}}
    \end{subfigure}
   \begin{subfigure}[b]{0.09\linewidth}
    \includegraphics[width=\linewidth]{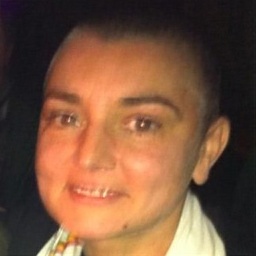}
    \caption{\cite{deblurganv2}}
    \end{subfigure}
  \begin{subfigure}[b]{0.09\linewidth}
    \includegraphics[width=\linewidth]{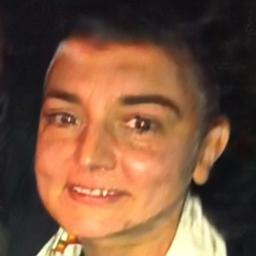}
    \caption{\cite{zhang2019deep}}
  \end{subfigure}
  \begin{subfigure}[b]{0.09\linewidth}
    \includegraphics[width=\linewidth]{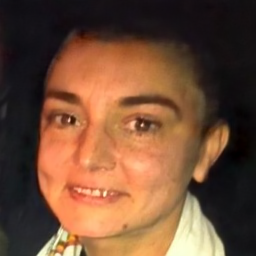}
    \caption{\cite{nah2017deep}}
  \end{subfigure}
  \begin{subfigure}[b]{0.09\linewidth}
    \includegraphics[width=\linewidth]{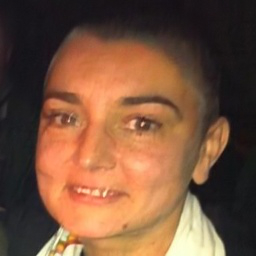}
    \caption{\cite{suin2020spatially}}
  \end{subfigure}
  \begin{subfigure}[b]{0.09\linewidth}
    \includegraphics[width=\linewidth]{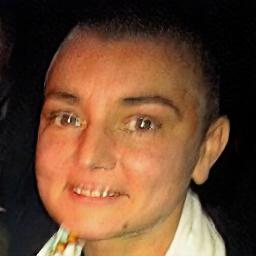}
    \caption{\cite{cyclegan}}
  \end{subfigure}
  \begin{subfigure}[b]{0.09\linewidth}
    \includegraphics[width=\linewidth]{Comparisons/unsupervised/face.jpg}
    \caption{\cite{lu2019unsupervised}}
  \end{subfigure}
  \begin{subfigure}[b]{0.09\linewidth}
    \includegraphics[width=\linewidth]{Comparisons/cyclegan/face.png}
    \caption{Ours}
  \end{subfigure}

  \caption{Visual comparisons with start of the art results on real blurred face images of \cite{real_blur}.}
  \label{fig:Visualcompar_real}
\end{figure*}

We used PSNR, NIQE and BRISQUE to provide quantitative comparisons with state of the art results. While PSNR requires ground truth or reference image, NIQE and BRISQUE do not require any reference image and can be calculated given a single image. A brief discussion of BRISQUE and NIQE is given below.

\textbf{BRISQUE} \cite{brisque} stands for Blind/Referenceless Image Spatial Quality Evaluator. BRISQUE uses scene statistics instead of distortion stats to calculate the naturalness of the given image. The low computational capacity of BRISQUE makes it well-suited for real-world applications. A lower BRISQUE score on an image indicates good perceptual quality, and its values range between 1-100.

\textbf{PIQE} \cite{piqe} stands for Perception-based Image Quality Evaluator. PIQE is a no-reference image metric that calculates the distortion present in the image based on block-level characteristics. PIQE estimates the quality of the image from perceptually significant portions rather than the whole image. Similar to BRISQUE, a lower score of PIQE indicates a better perceptual score, and its value ranges between 1-100.

% \begin{align*}
%     V = \text{Concat}(U^a_{X}, U_{2X})
% \end{align*}
\begin{figure*}[t!]
% \hskip-0.7cm
  \centering
 \begin{subfigure}[b]{0.15\linewidth}
    \includegraphics[width=\linewidth]{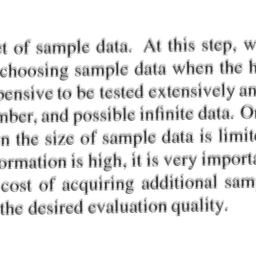}
    \caption{Blurred}
  \end{subfigure}
  \begin{subfigure}[b]{0.15\linewidth}
    \includegraphics[width=\linewidth]{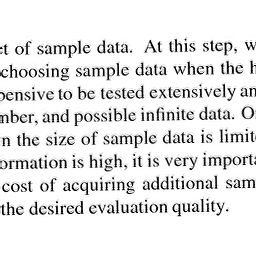}
    \caption{\cite{xu2013unnaturall0}}
    \end{subfigure}
    \begin{subfigure}[b]{0.15\linewidth}
    \includegraphics[width=\linewidth]{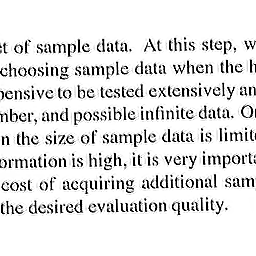}
    \caption{\cite{textdeblurring}}
    \end{subfigure}
  \begin{subfigure}[b]{0.15\linewidth}
    \includegraphics[width=\linewidth]{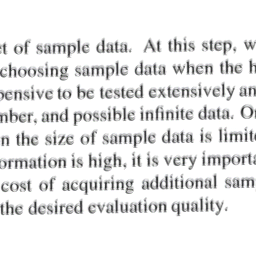}
    \caption{\cite{kupyn2018deblurgan}}
    \end{subfigure}
   \begin{subfigure}[b]{0.15\linewidth}
    \includegraphics[width=\linewidth]{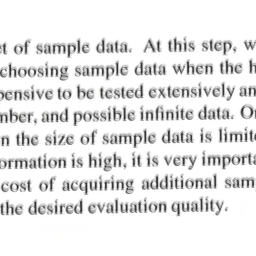}
    \caption{\cite{deblurganv2}}
    \end{subfigure}
  \begin{subfigure}[b]{0.15\linewidth}
    \includegraphics[width=\linewidth]{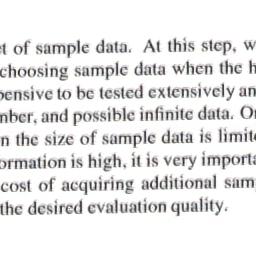}
    \caption{\cite{zhang2019deep}}
  \end{subfigure}\\
  \begin{subfigure}[b]{0.15\linewidth}
    \includegraphics[width=\linewidth]{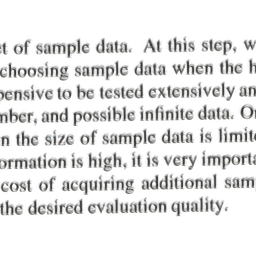}
    \caption{\cite{nah2017deep}}
  \end{subfigure}
  \begin{subfigure}[b]{0.15\linewidth}
    \includegraphics[width=\linewidth]{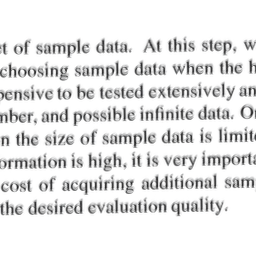}
    \caption{\cite{suin2020spatially}}
  \end{subfigure}
  \begin{subfigure}[b]{0.15\linewidth}
    \includegraphics[width=\linewidth]{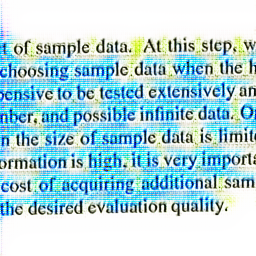}
    \caption{\cite{cyclegan}}
  \end{subfigure}
  \begin{subfigure}[b]{0.15\linewidth}
    \includegraphics[width=\linewidth]{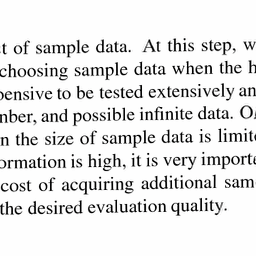}
    \caption{\cite{lu2019unsupervised}}
  \end{subfigure}
  \begin{subfigure}[b]{0.15\linewidth}
    \includegraphics[width=\linewidth]{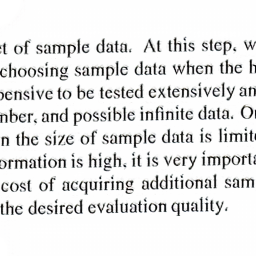}
    \caption{Ours}
  \end{subfigure}
  \begin{subfigure}[b]{0.15\linewidth}
    \includegraphics[width=\linewidth]{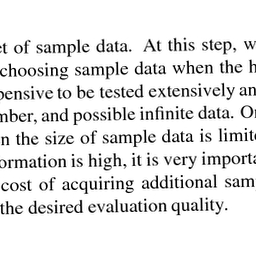}
    \caption{Sharp}
  \end{subfigure}
  \caption{Visual comparisons with start of the art results on text dataset \cite{text}.}
  \label{fig:Visual_comp_text}
\end{figure*}
\begin{table}[!htb]
    % \caption{Global caption}
    % \begin{minipage}{.5\linewidth}
      \centering

\caption{Quantitative comparisons with state of the art methods on the face and text dataset.}
\hskip-0.7cm
\begin{tabular}{|c|c|c|c|c|c|c|}
    \hline
     Method & \multicolumn{3}{|c|}{Face datset} &  \multicolumn{3}{|c|}{Text dataset} \\\cline{2-7}
    & \textit{brisque} & \textit{piqe} & PSNR & \textit{brisque} & \textit{piqe} & PSNR \\
     \hline
     \cite{textdeblurring} & \xmark & \xmark & \xmark & 42.35 & 76.06 & 17.04\\
     \hline
     \cite{xu2013unnaturall0} & 36.82 & 55.41 & 18.07 & 45.15 & 77.87 & 15.30 \\
     \hline
     \cite{kupyn2018deblurgan} &43.54 & 57.32& 18.61 &  47.34 & 80.43 & 17.67 \\
     
     \hline
     \cite{deblurganv2} & 44.36 & 57.78 &\textbf{19.34} & 46.58 & 80.76 & 
17.90 \\
     \hline
     \cite{zhang2019deep} &48.25 & 71.0 &19.00 &43.92  & 76.23 &  17.48 \\
     \hline
     \cite{nah2017deep} & 47.88 & 77.73 & 18.62 &46.69  &81.33 & 17.84 \\
     \hline
     \cite{suin2020spatially} & 44.77 & 66.09 & 19.21 & 46.46 &81.74  & 
\textbf{18.97}\\
     \hline
     \cite{cyclegan} &  31.07 &42.83 &  18.68 & 48.32 & 80.32& 14.56\\
     \hline
      \cite{lu2019unsupervised} & 29.97 & 45.03 & 19.05 &47.19 & 79.94& 18.49\\
     \hline
     Ours &  \textbf{25.52} & \textbf{35.93} &19.24 & \textbf{39.64} & \textbf{74.05} & 18.68 \\
     \hline
    \end{tabular}
\label{Quantitative analysis}

\end{table}

% \textbf{Net7:} (Table \ref{ablationTable}, row 7 and Fig. \ref{fig:Ablation_study1} (h)): Similar to Net6, spatial attention \cite{spatialattention} is applied instead of channel attention on previous scale features and the resultant concatenated features are passed through the decoder. As observed, due to pixel-wise attention, the previous scale feature maps are better attended, further helping the deblurring quality.

% \textbf{Net8:} (Table \ref{ablationTable}, row 8 and Fig. \ref{fig:Ablation_study1} (i)): For both Net6 and Net7, the feature maps of the present scale do not play any role in attending to previous scale features. Motivated by this, we propose to use SAAM. SAAM attends to feature maps from the previous scale as a function of the present scale. As can be seen, Net8 gives improved deblurring performance compared to previous networks. 
\subsection{Competing methods}
The results of our model are compared with conventional methods \cite{textdeblurring, xu2013unnaturall0}, supervised methods \cite{zhang2019deep, kupyn2018deblurgan, deblurganv2, nah2017deep, suin2020spatially} and unsupervised methods \cite{cyclegan, lu2019unsupervised}. Among conventional methods, \cite{textdeblurring} is a text deblurring method, while \cite{xu2013unnaturall0} is a generic deblurring algorithm. In CNN based methods, \cite{lu2019unsupervised, cyclegan} are domain-specific methods and \cite{suin2020spatially, zhang2019deep, deblurganv2, kupyn2018deblurgan} are natural scene deblurring methods. For conventional methods, we ran the codes with default parameters provided by authors, while for CNN methods, we used the pretrained models provided by authors except for CycleGAN \cite{cyclegan}. We used the official code provided by authors to retrain the CycleGAN\cite{cyclegan} on face and text training datasets.  

\subsection{Comparisons}
\textbf{Test dataset results:} Fig. \ref{fig:Visualcompar_face} and Fig. \ref{fig:Visual_comp_text} shows visual comparisons, while Table \ref{Quantitative analysis} illustrates quantitative comparisons with competing methods on the faces and text test set (described in Sec. 4.1). Our method outperforms conventional and unsupervised methods on all three metrics. Compared with supervised methods, our method performs comparably on the PSNR metric while giving superior performance on no-reference metrics. From Fig. \ref{fig:Visualcompar_face} and \ref{fig:Visual_comp_text},  we can see that \cite{xu2013unnaturall0} over blurs the image at specific regions and neglects the other portions, while the deblurring quality is poor in \cite{deblurganv2, kupyn2018deblurgan}. Among supervised methods, \cite{zhang2019deep, suin, nah2017deep} gives comparably good results due to recurrent structure but fails to deblur specific portions. In unsupervised methods, CycleGAN\cite{cyclegan} induces artifacts in the restored image (Fig. \ref{fig:Visual_comp_text} (i)) while \cite{lu2019unsupervised} fails to properly recover the latent image when encountered by complex blur (Fig. \ref{fig:Visualcompar_face} (i).

\textbf{Real dataset results:} We cropped nine face images from the real world blurry images provided by \cite{real_blur} and the corresponding visual comparisons are shown in Fig. \ref{fig:Visualcompar_real}. Consistent with test dataset results, \cite{xu2013unnaturall0} tends to over blur some portions of the image while \cite{deblurganv2, kupyn2018deblurgan} leave most of the portions to remain blurred. 
\cite{zhang2019deep, suin, nah2017deep} gives good results on the first image due to scale recurrent nature; however, some second image portions remain blurred. In unsupervised methods, \cite{cyclegan} fails to recover the clean domain while \cite{lu2019unsupervised} struggles to restore the clean image when a large amount of blur is present.  Compared to the above methods, our methods give superior performance while handling blurred faces of the test dataset and real-world face images. We provide further analysis of the real-world dataset using no reference metrics in the supplementary material.

\section{Conclusions}
In this paper, we proposed a multi-scale unsupervised network for deblurring domain-specific data. We used a coarse-to-fine approach to stabilize GAN training and a scale adaptive attention module (SAAM) to aid relevant information flow across scales. Ablation studies show the importance of using our multi-scale approach in conjunction with SAAM. Qualitative and quantitative comparisons show that our methods perform on par with supervised methods while outperforming conventional and unsupervised methods.
{\small
\bibliographystyle{ieee}
\bibliography{egbib}
}

\end{document}